\documentclass{article}

\usepackage{microtype}
\usepackage{graphicx}
\usepackage{subfigure}
\usepackage{booktabs} 
\usepackage{stfloats}

\usepackage{hyperref}

\usepackage[accepted]{icml2025}

\usepackage{amsmath}
\usepackage{amssymb}
\usepackage{mathtools}
\usepackage{amsthm}

\usepackage[capitalize,noabbrev]{cleveref}

\theoremstyle{plain}

\theoremstyle{definition}

\theoremstyle{remark}

\usepackage[textsize=tiny]{todonotes}

\icmltitlerunning{Submission and Formatting Instructions for ICML 2025}

\begin{document}

\twocolumn[
\icmltitle{Time-series Forecast for Indoor Zone Air Temperature with Long Horizons: A Case Study with Sensor-based Data from a Smart Building}



\icmlsetsymbol{equal}{*}

\begin{icmlauthorlist}
\icmlauthor{Liping Sun}{equal,sch2}
\icmlauthor{Yucheng Guo}{equal,sch1}
 \icmlauthor{Siliang Lu}{comp1}
 \icmlauthor{Zhenzhen Li}{comp2}
\end{icmlauthorlist}

\icmlaffiliation{comp1}{Bosch Center for Artificial Intelligence,  Shanghai, China}
\icmlaffiliation{comp2}{Bosch Center for Artificial Intelligence,  Pittsburgh, USA}
\icmlaffiliation{sch1}{College of Architecture and Urban Planning, Tongji University, Shanghai, China}
\icmlaffiliation{sch2}{Department of Statistics, Iowa State University, Ames, USA}
\

\icmlcorrespondingauthor{Siliang Lu}{siliang.lu@cn.bosch.com}

\icmlkeywords{Hybrid Model; Long-term Prediction; Zone Air Temperature; VAV System }

\vskip 0.3in
]



\printAffiliationsAndNotice{\icmlEqualContribution}  

\begin{abstract}
  With the press of global climate change, extreme weather and sudden weather changes are becoming increasingly common. To maintain a comfortable indoor environment and minimize the contribution of the building to climate change as much as possible, higher requirements are placed on the operation and control of HVAC systems, e.g., more energy-efficient and flexible to response to the rapid change of weather. This places demands on the rapid modeling and prediction of zone air temperatures of buildings. Compared to the traditional simulation-based approach such as EnergyPlus and DOE2, a hybrid approach combined physics and data-driven is more suitable. Recently, the availability of high-quality datasets and algorithmic breakthroughs have driven a considerable amount of work in this field. However, in the niche of short- and long-term predictions, there are still some gaps in existing research. This paper aims to develop a time series forecast model to predict the zone air temperature in a building located in America on a 2-week horizon.  The findings could be further improved to support intelligent control and operation of HVAC systems (i.e. demand flexibility) and could also be used as hybrid building energy modeling.

\end{abstract}

\section{Introduction}
\label{submission}

In order to address the impacts of extreme weather events and global climate change. HVAC system plays an important role in the field of indoor environment control and carbon emission reduction. Real-time dynamic control of the system is a prerequisite for achieving the above two objectives, which requires fast and accurate predictions of room temperature, so that the setpoint of the air system and the water system can be set at a suitable value.

Some simulation tools such as EnergyPlus and DOE2 with their derivative software could be applied to compute the zone air temperature in abstract thermodynamic networks. However, dynamic simulation requires high computing power and lacks flexibility when some input parameters are changed \cite{reinhart2016urban} \cite{nutkiewicz2021exploring}. In contrast, for real-time rapid prediction, data-driven approaches offer better performance with high computational efficiency and fast adjustment of the input feature. Recently, some research has reported the precision and efficiency of this approach, such as Godinho et al. \cite{Godinho2021app} and Hu et al. \cite{Hu2023adv}. However, related works normally utilize a considerable number of features as input included exogenous variants, which improve difficulty in obtaining reliable datasets and may hinder the understanding of model predictions. 
 
To address the above issues, hybrid modeling (or gray-box modeling) is considered a potential solution. Therefore, it could take advantage of both the simulation-based or physics-based approaches as well as the data-driven approach \cite{nutkiewicz2018data}. Typically, the hybrid model combines statistical techniques to reduce the number of physical building characteristics that need to be input. There are several methods to achieve this goal. A number of work aims to develop a simulation-based tool equipped with a comprehensive parameter database Based on extensive statistical data to reduce the required input information, such as SimStadt \cite{nouvel2015simstadt}, City Energy Analyst \cite{fonseca2016city} and TEASER \cite{remmen2018teaser}.  Another common method is to calibrate the Resistor-Capacitor (RC) model using optimization algorithms to predict indoor temperature \cite{wei2022data} \cite{li2017development}, thermal load \cite{ogunsola2014development} \cite{ogunsola2015application} , and other factors such as solar heat gain \cite{omar2017self} and peak load \cite{vivian2017evaluation}. Furthermore, another possible implementation method is to apply the simulation approach in some zones and a data-driven approach in other zones\cite{FOUCQUIER2013272}. In addition, with the help of explainable tools, the model based on data-driven approach could be understood the impact of different input features, which is not an absolute "black-box" model. This kind of model could also be considered as a hybrid model because the explanation is usually based on physics principles \cite{fan2019novel} \cite{mouakher2022expect}.   

As mentioned above, a considerable number of research studies have attempted to explain the relationships between input features and responses to the building system by post hoc interpretability \cite{lipton2018mythos} with the help of tools such as SHapley Additive exPlanations  (SHAP) \cite{lundberg2017unified} and Local Interpretable Model-Agnostic Explanations  (LIME) \cite{ribeiro2016should}. This can enhance users' understanding and trust in the model's principles. However, the relationships between the parameters themselves may be even more important, as this could simplify data collection as much as possible, thus reducing the cost of data sampling and improving the transferability of research results.

In the above context, this paper aims to develop a time series forecast model with explainable features for zone air temperature for intelligent building control such as agent-based reinforcement learning. The two potential highlights are as follows:

    \textbf{Long-term time-series auto-regressive prediction}: Long-term prediction with time series auto-regression and thermodynamic principles for zone air temperature to optimize the HVAC system control strategy.
    
    \textbf{Spatial-temporal prediction for different zones}: As the impacts of features on zone air temperature forecast differ from one location to another (i.e. perimeter zone vs core zone), the paper investigates on how to improve the accuracy of the prediction model by incorporating such spatial information.

\section{Problem understanding}

As mentioned in the Introduction, the goal of the article is to develop a model in which, given the exogenous validation information, it can predict the zone air temperature. Furthermore, with this amount of data to model and understand important building dynamics, it can be used for downstream tasks such as determining thermal comfort and reducing energy consumption for the cooling and heating systems of smart buildings.

Data were collected from a smart building that is equipped with a typical VAV system\cite{goldfeder2025smartbuildingscontrolsuite}. The main contents of the data set include the following elements:

1. The observation values and action values of the HVAC system in this building.

2. The outside whether data.

3. Information about the HVAC system/devices.

4. The floor plan of the building and the distribution of the devices.

\begin{figure}[h]
\vskip 0.2in
\begin{center}
\centerline{\includegraphics[width=\columnwidth]{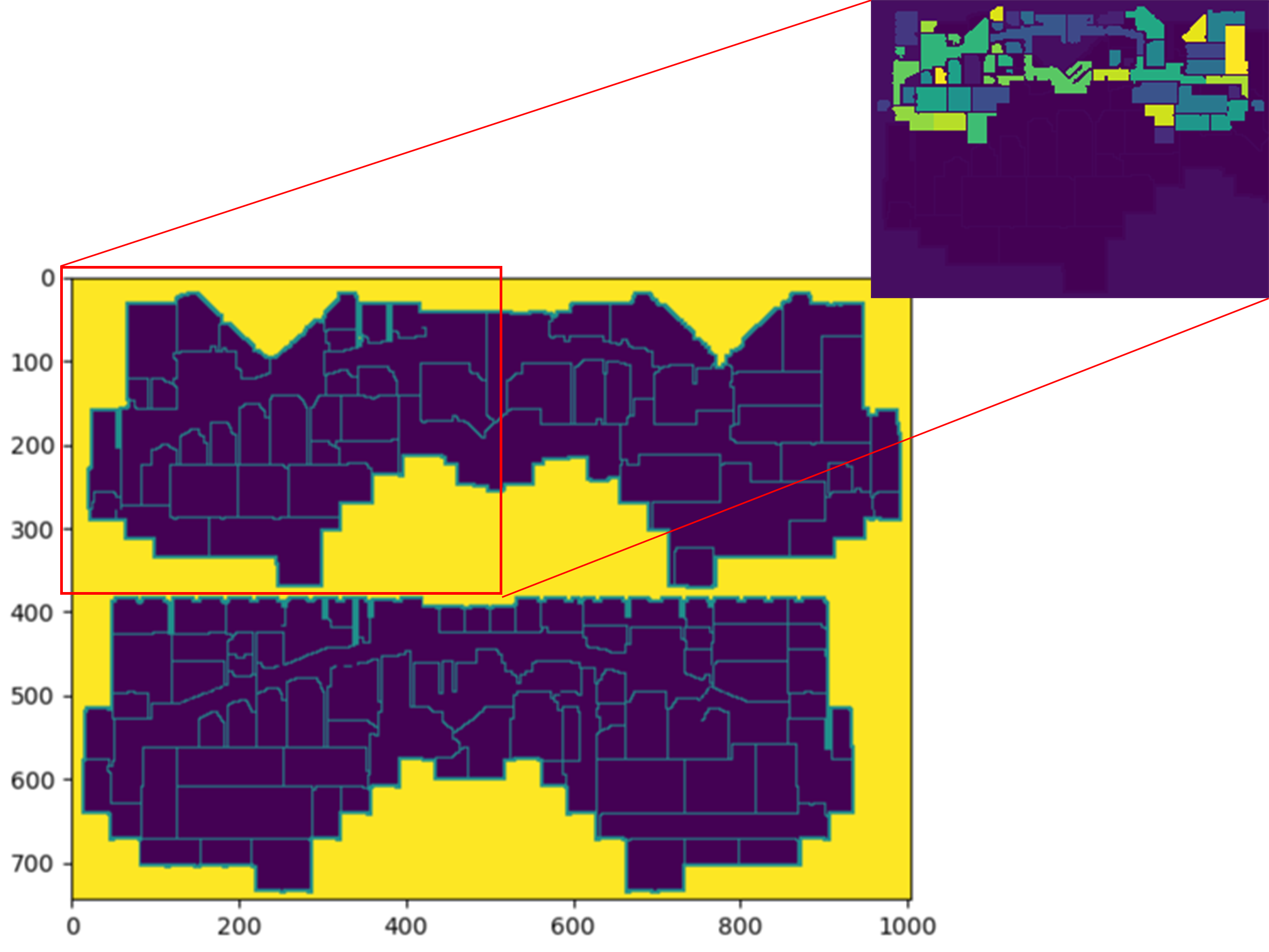}}
\caption{Floorplan of the case building}
\label{floorplan of the case building}
\end{center}
\vskip -0.2in
\end{figure}

For the contents of 1 and 2, the data include five datasets from January 2022 to June 2024, e.g., 2022a, 2022b, 2023a, 2023b, 2024a. The data in each single year were divided into 2 parts, data in the first 6 months will be used as a train dataset, and the last 6 months will be used as validation dataset. For the contents of 3, the ids, names, types, and monitored values of the devices were listed, and for the contents of 4, the floor plans of the building and the distribution of the devices were used as a two-dimensional matrix to illustrate the horizontal coordinates. The details are shown in \cref{Content}. Meanwhile, the floor plan and the device distribution are shown in \cref{floorplan of the case building}. As shown in the figure, it is a typical all-air variable air volume (VAV) system consisting of a chilled and cooling water system, an air handling unit, and multiple VAV units in each thermal zone. 

\begin{table*}[ht]
\caption{The Content of the Dataset}
\label{Content}
\vskip 0.05in
\begin{center}
\begin{small}
\begin{sc}
\begin{tabular}{p{4cm}p{12cm}}
\toprule
Data Type & Description \\
\midrule
Observation Values (All Devices) & 
Supply air flow rate set point, supply air flow rate sensor, zone air cooling temperature set point, zone air heating temperature set point, zone air temperature sensor \\
Observation Values (Partial Devices) & 
Zone air CO\textsubscript{2} concentration set point, zone air CO\textsubscript{2} concentration sensor, discharge air temperature set point, outside air flow rate set point, outside air flow rate sensor, discharge air temperature sensor, mixed air temperature set point, mixed air temperature sensor, etc. \\
Action Values of Devices & 
Supply air temperature set point, supply water temperature set point \\
Outside Weather Data & 
Outside air wet bulb temperature sensor, outside air temperature sensor, outside air specific enthalpy sensor, outside air relative humidity sensor, outside air dew point temperature sensor \\
Device Metadata & 
Device ID, device namespace, device type, data recorded by the device \\
\bottomrule
\end{tabular}
\end{sc}
\end{small}
\end{center}
\vskip -0.1in
\end{table*}

As shown in \cref{Content}, the size of this data set is relatively large with more than 2GB in total.  There are more than 1,000 variables, including outdoor air temperature, power, etc. Thus, preprocessing the data required domain knowledge to identify several key features as input for the model.  As a result, we selected the input features that include outdoor air temperature, actual air temperature, and flow rate of the zone, heating and cooling set points of the zone for each forecast of air temperature of the zone, as shown in \cref{data structure} below.

\begin{figure}[ht]
\vskip 0.2in
\begin{center}
\centerline{\includegraphics[width=\columnwidth]{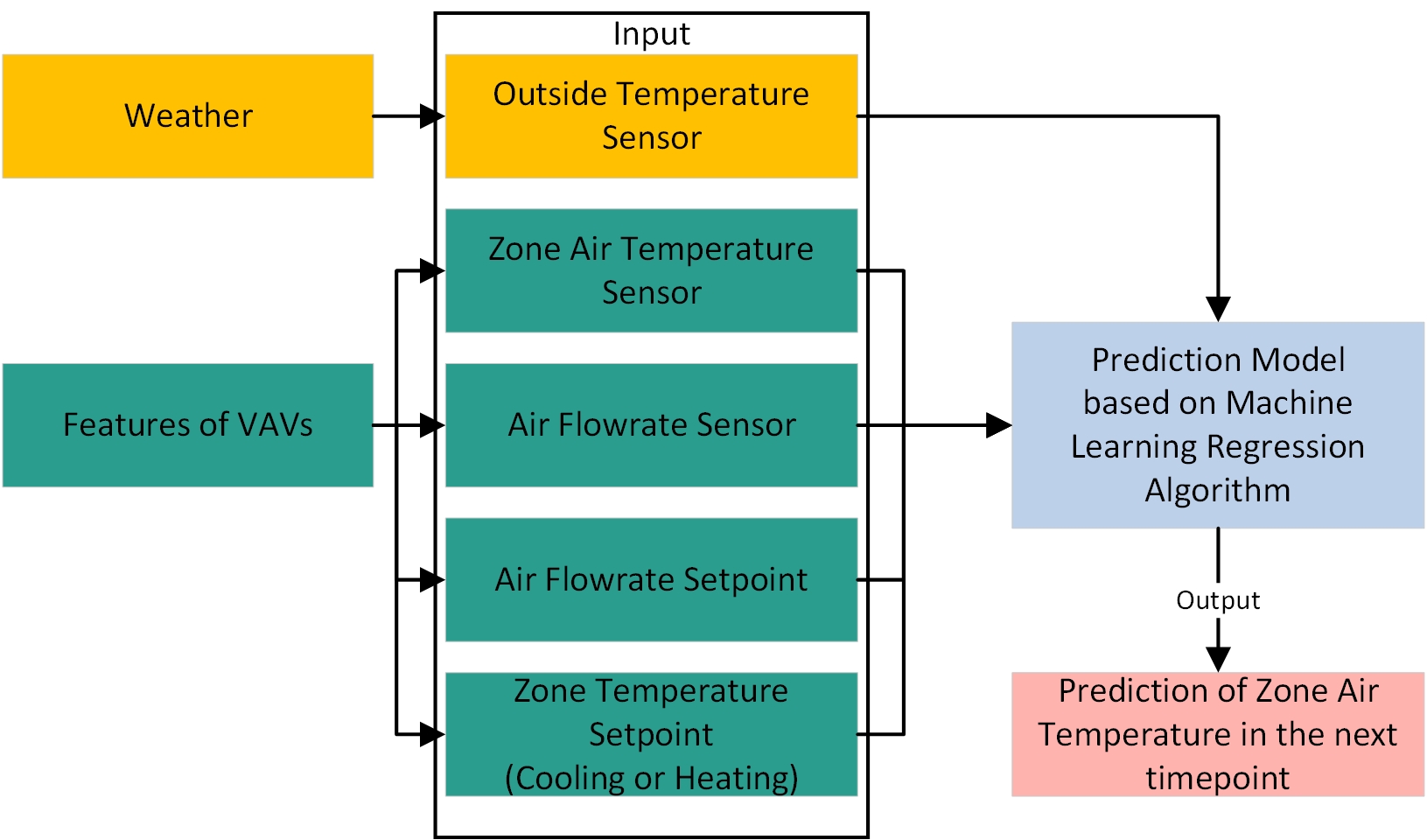}}
\caption{Input and Output Used for the Model}
\label{data structure}
\end{center}
\vskip -0.2in
\end{figure}

For the first phase, we have made the following assumption: temperature changes in each area corresponding to each VAV are independent of each other. Under this assumption, the model could be trained with the data of only each individual VAV. However, for the core and perimeter zones, due to differences in heat transfer levels between the indoor and outdoor environment, the impact of outdoor temperatures on indoor zone air temperatures is undoubtedly different. Therefore, in order to reflect such differences, in the second phase, model correction and validation are necessary. In addition, since the time step in the dataset is 5 minutes, frequent predictions are not necessary as adjustments cannot achieve ideal control results considering the thermal mass effects in building envelopes. Therefore, we further down-sampled the data and developed models with a timestep of 15 minutes and 1 hour, respectively.

Therefore, based on the above analysis using prior/domain knowledge in the fields of HVAC and building physics, combined with the characteristics of the dataset, we aim to develop a time-series forecast model for zone air temperature prediction based on data from each individual VAV box first. Furthermore, the impact of spatial location on the prediction results will be verified and corrections will be established based on the location of the VAV box.

\section{Methodology}
To develop the model, we split the data into train data and validation data. For the data in 2022 and 2023, the data in the first 6 months of the year will be used as train dataset and in the last 6 months will be used as validation dataset. To further validate the model's performance, we used data from the first half of 2024 as the test set. Meanwhile, in order to achieve predictions across different time spans, we sliced the data according to a duration of 2 weeks.

The zone air temperature forecast is developed with a hybrid data-driven and thermodynamic model. After preprocessing including feature selection, missing data imputation, and outlier removal, the input features of the model include outside air temperature and actual zone air temperature, actual zone air temperature setpoint, actual supply pressure setpoint of the last n (i = 1,2,..n) weeks, and the output is the predicted temperature of the next n weeks (i = 1,2..n). The forecast model diagram is shown below:
\begin{figure}[ht]
\vskip 0.2in
\begin{center}
\centerline{\includegraphics[width=\columnwidth]{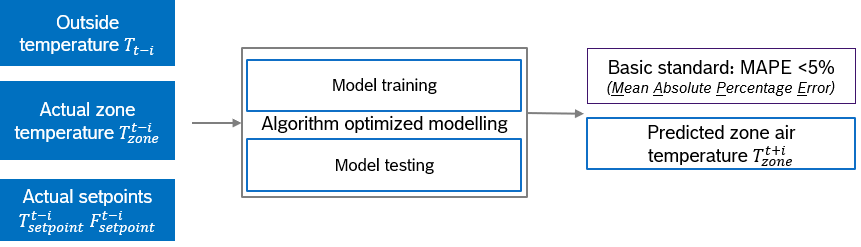}}
\caption{Data-driven thermodynamics model }
\label{temp}
\end{center}
\vskip -0.2in
\end{figure}

Moreover, for time-series forecast models, a hierarchical prediction module is developed, and the best model is selected with a model selection algorithm called time-series cross-validation. As mentioned above, the paper uses various regression algorithms, including random forecast regression, Gaussian process regression, Adaboost regression, gradient boosting regression, and XGBoost regression. All different regressions are developed with multivariate auto-regressive approach. In addition, mean absolute error (MAE) and root mean square error (RMSE) are used to evaluate the performance of the forecast models.

\section{Post-launch performances}
By comparing performance of the different models, a time-series prediction model based on XGBoost combined with self-regression is developed to predict the zone air temperature with a horizon of 2 weeks.

\begin{figure}[ht]
\vskip 0.2in
\begin{center}
\centerline{\includegraphics[width=\columnwidth]{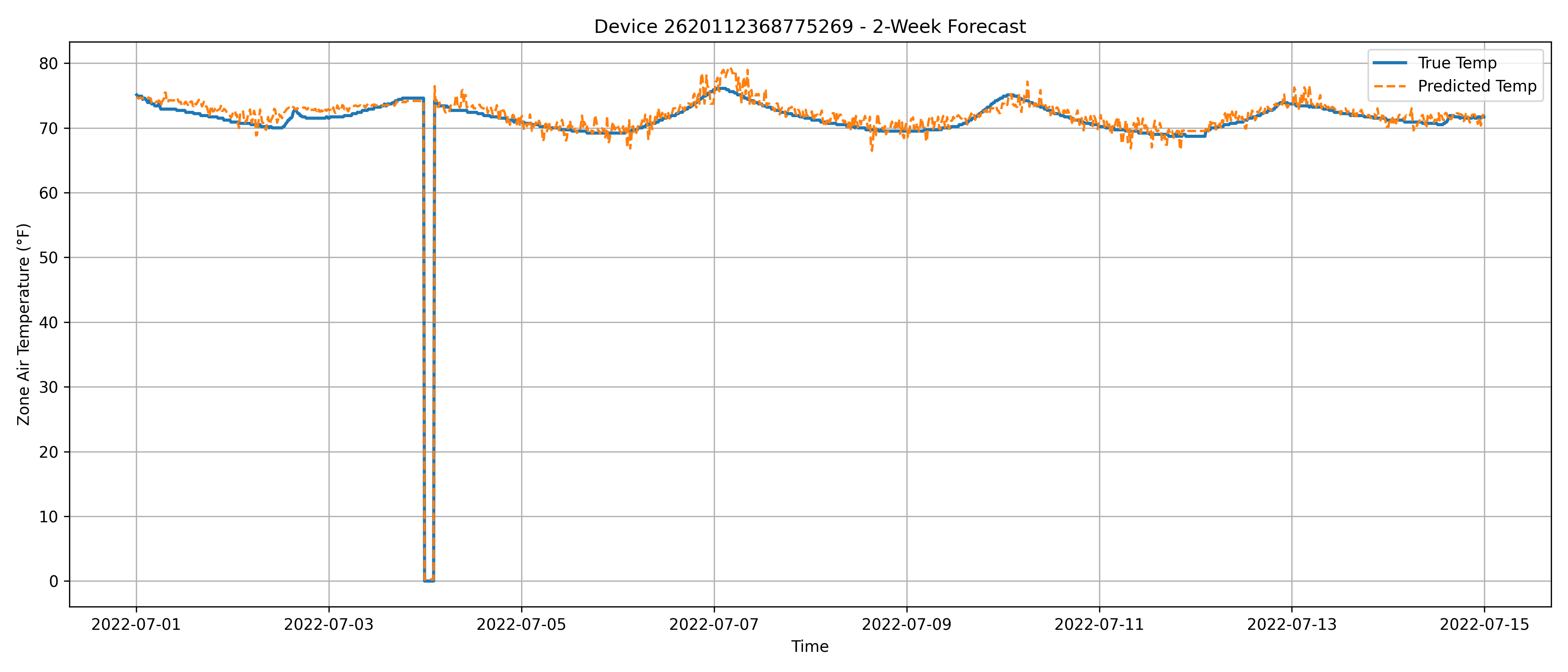}}
\caption{Post-launch performance of indoor air temperature changes for the 1st VAV }
\label{indoor1}
\end{center}
\vskip -0.2in
\end{figure}

\begin{figure}[ht]
\vskip 0.2in
\begin{center}
\centerline{\includegraphics[width=\columnwidth]{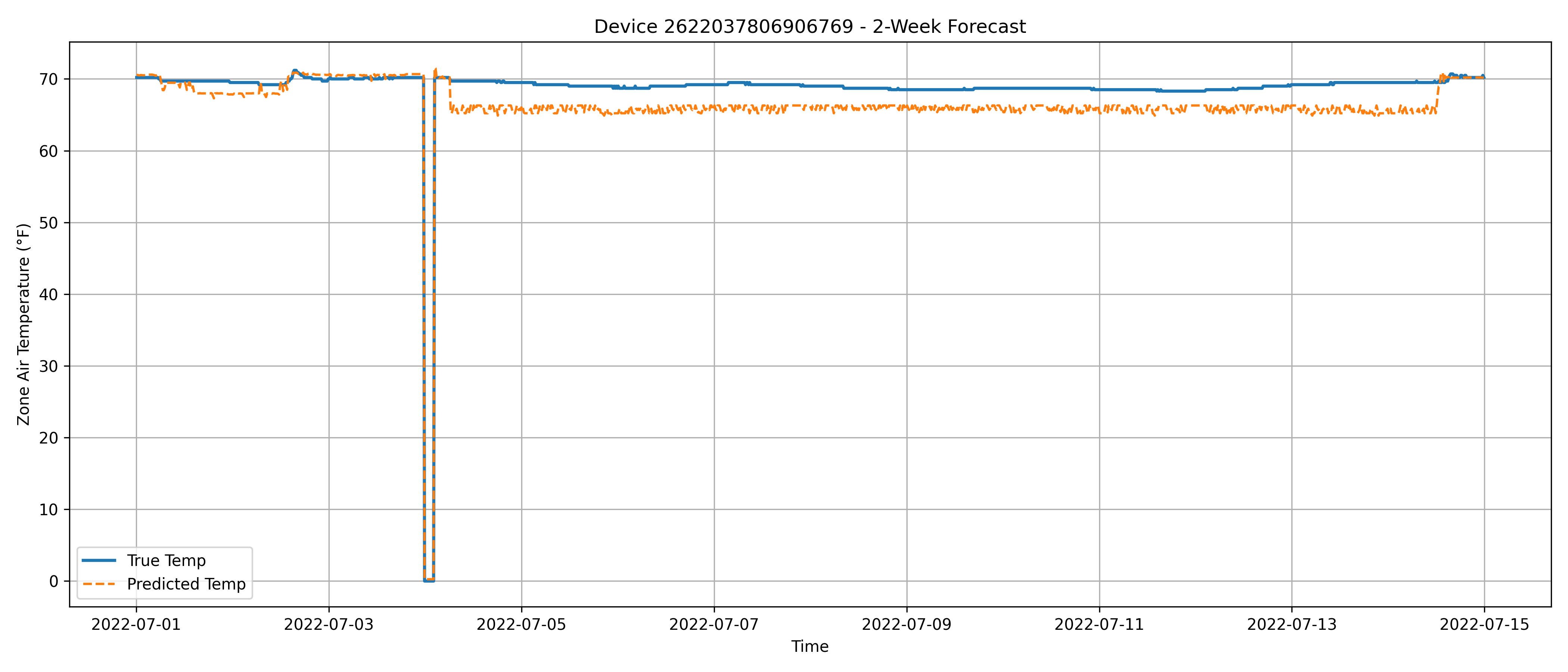}}
\caption{Post-launch performance of indoor air temperature changes for the 2nd VAV }
\label{indoor2}
\end{center}
\vskip -0.2in
\end{figure}

\begin{figure}[ht]
\vskip 0.2in
\begin{center}
\centerline{\includegraphics[width=\columnwidth]{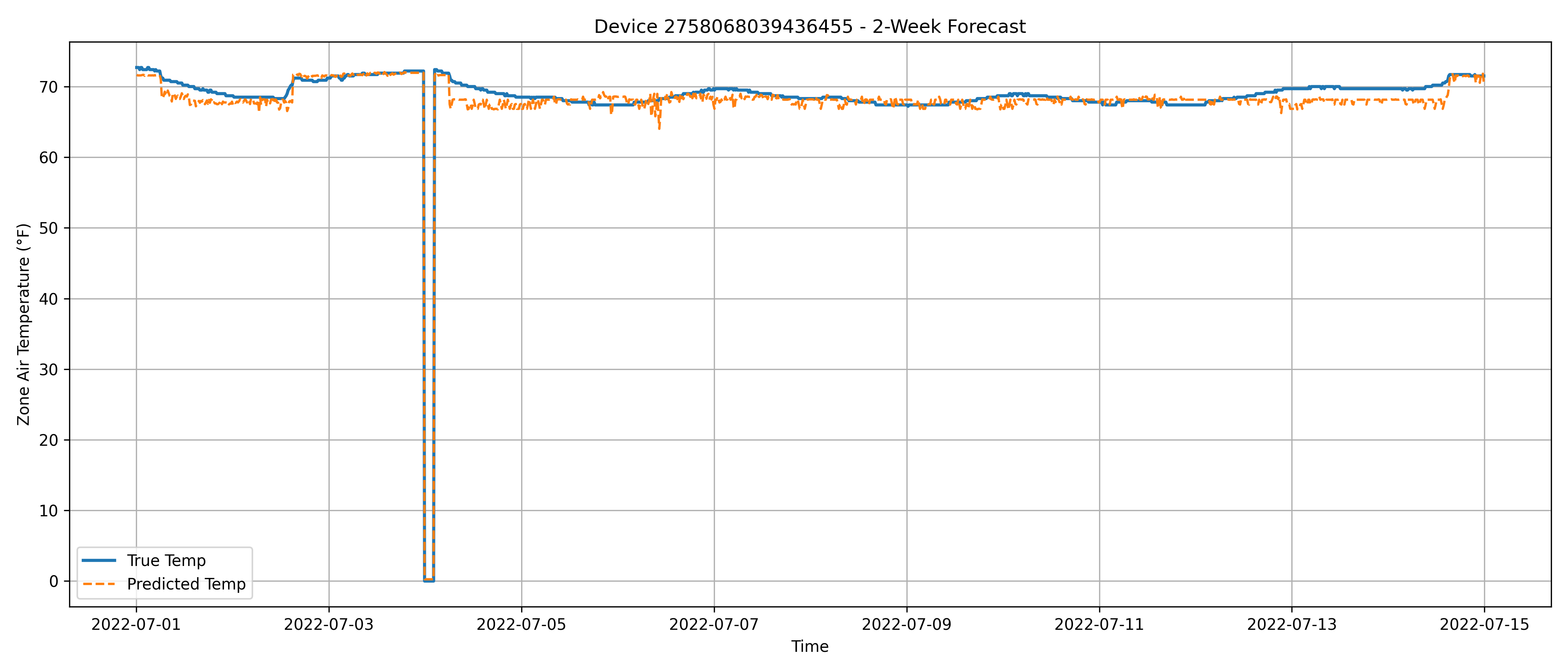}}
\caption{Post-launch performance of indoor air temperature changes for the 3rd VAV }
\label{indoor3}
\end{center}
\vskip -0.2in
\end{figure}

As a result, the average MAE is 4.20 and the average RMSE is 4.8, indicating that the air temperature error is approximately $4^{\circ}\mathrm{F}$. Therefore, in the second phase a more accurate zone air temperature prediction could be achieved. In addition, we selected 3 VAVs located in different positions to illustrate the performance of the prediction model. The first is on the 2nd floor which is named as 'VAV RH 2-2-50' with No. 2620112368775269. The second is located in the north room on the first floor which is named as 'VAV RH 1-1-19' with No. 2622037806906769. The third is located in the south room on the first floor which is named as 'VAV CO 1-1-51' with No. 2758068039436455. The predicted  zone air temperature for the above 3 VAVs in 2 weeks is shown in \cref{indoor1}, \cref{indoor2}, \cref{indoor3}. As shown in figure, the zone air temperature  has fluctuated between $65^{\circ}\mathrm{F}$ and $75\,^{\circ}\mathrm{F}$. 

For the above VAV boxes, the performances differ from each other. This may be due to two potential reasons. Firstly, since the model is independent of spatial location, this may affect the impact of solar radiation, which is not an input parameter of the model. The second reason may be that the devices have different functions. The above two points are potential paths for further optimization and improvement of the model in the future.

\section{Potential application}
For smart building systems, with a long-horizontal time series forecast model, real-time optimization of the HVAC system could be achieved for energy and cost savings. Here is one of the potential applications.

Even if local policies support hourly flexible tariffs according to seasons, there is no response to the dynamic shift of the peak valley of power for most commercial buildings, which causes cost and energy waste. Fortunately, due to the inherent flexibility of the building cooling loads from the thermal mass characteristics of various building envelopes (i.e., wall, ceiling, etc.), a more comfortable and energy-efficient environment could be achieved by energy flexibility optimization.

The overall energy flexibility optimization case study can be divided into mainly two parts: zone air temperature forecast, and temperature \& pressure setpoint control. The utility rate there has flexible tariffs, and the optimized actions can be transmitted in real time to the machine with an industrial protocol such as Modbus or BACNet.

\begin{figure}[ht]
\vskip 0.2in
\begin{center}
\centerline{\includegraphics[width=\columnwidth]{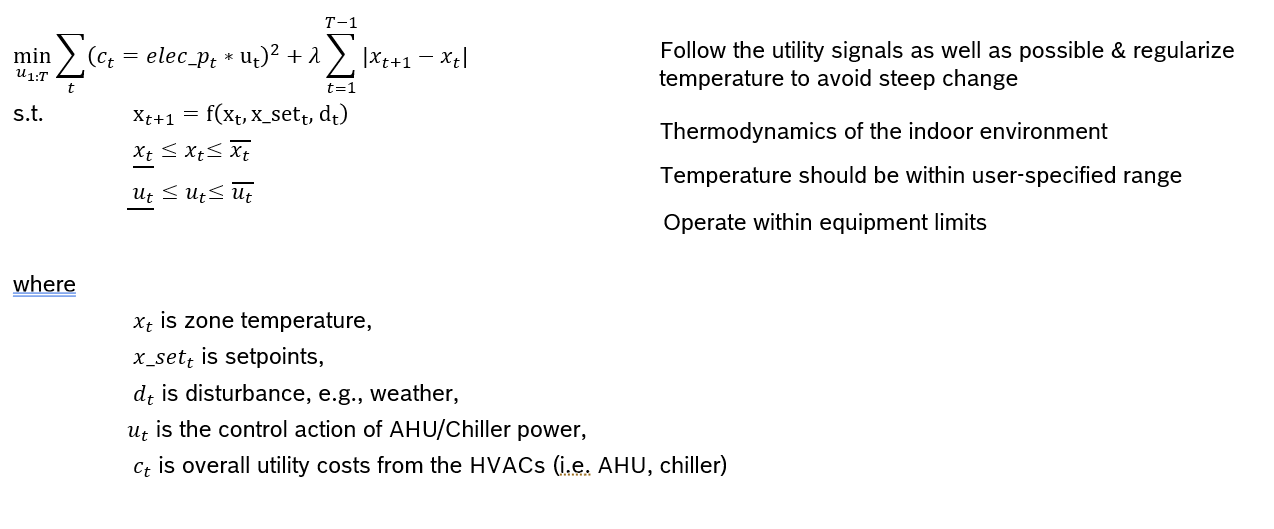}}
\caption{Energy flexibility Optimization model}
\label{opt}
\end{center}
\vskip -0.2in
\end{figure}

With the prediction model, for each timestep, the setpoint controls are based on the convex optimization model using the data-driven forecast model and flexible tariffs. The optimization model has the following formula in figure \ref{opt}.

 The objective function is to minimize both the overall costs and the deviation between the sequential indoor air temperature to follow the tariff signal as much as possible \& regularize temperature to change slowly. Since it is assumed that the AHU supply pressure differential setpoint has a linear correlation with the AHU power following \( P \propto \Delta p \), the pressure differential setpoint can be derived from the AHU power. Moreover, since the AHU supply temperature setpoint affects the chiller power, it can also be derived from another control variable of the chiller power. A simplified calculation formula is used to calculate the change in the chiller power.
\[
Q_{\text{cool demand}} = \sum_{i=1}^{n} \dot{m}_{\text{air},i} \cdot c_{p,\text{air}} \cdot (T_{\text{zone},i} - T_{\text{zone setpoint},i})
\]
\[
Q_{\text{cool supply}} = \dot{m}_{\text{water}} \cdot c_{p,\text{water}} \cdot (T_{\text{return}} - T_{\text{supply}})
\]
\[
P_{\text{chiller}} =  \frac{Q_{\text{cool supply}}}{\text{COP}}
\]
\[
Q_{\text{cool demand}} =  Q_{\text{cool supply}}
\]

where the COP is calculated in real time \cite{lu2023real}, 
\( Q_{\text{cool demand}} \) represents the indoor cooling demand of the AHU, 
and \( Q_{\text{cool supply}} \) denotes the cooling supplied by the chiller to the AHU. It is assumed that there is only one AHU unit in the building.

\begin{figure}[ht]
\vskip 0.2in
\begin{center}
\centerline{\includegraphics[width=\columnwidth]{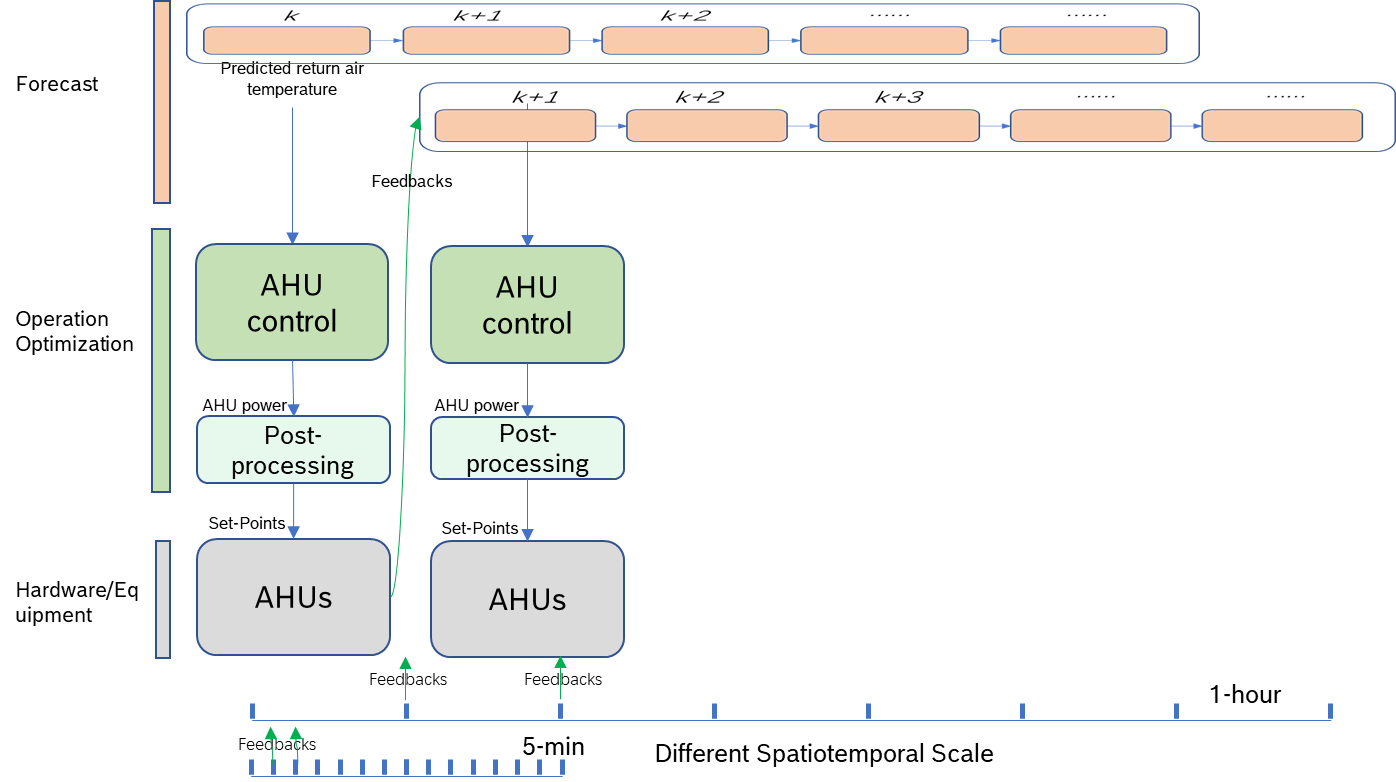}}
\caption{Data-driven model predictive control diagram }
\label{mpc}
\end{center}
\vskip -0.2in
\end{figure}

In total, with the forecast model and the optimization model, data-driven model predictive control is used to continuously control the setpoints optimally shown in \ref{mpc}. In addition to 1-hour AHU operation according to tariffs and predictions, predictive control of the 5-minute model with VAV boxes ensures further energy savings while still maintaining comfortable indoor air temperature in the smart building. 



\section{Conclusion}
This paper aims to propose a time-series forecast model for zone air temperature with long horizon. In addition, flexibility optimization of HVAC control strategies using the predicted indoor thermodynamics forecast model and data-driven model predictive control could be a potential downstream application. As a result, an ensemble forecast model for 2-week zone air temperature is developed and will take into account the location and orientation of each thermal zone in the second phase. Moreover, for the potential application, there is high scaling potential with cloud solutions for the data-driven model predictive control system in all commercial and industrial buildings with all related devices (i.e. batteries, thermal storage, PV panels) in the demand-supply energy system.

\section*{Ackownledgements}
We would like to thank the organizers of CO-BUILD workshop, and also the help from the PhD student Judah Goldfeder at Columbia University.

\section*{Impact Statement}
`This paper presents work whose goal is to advance the field of Machine Learning. There are many potential societal consequences of our work, none of which we feel must be specifically highlighted here. In the future, we would like to further optimize the model for accuracy.



\bibliography{example_paper}
\bibliographystyle{icml2025}

\newpage
\appendix
\onecolumn



\end{document}